\documentclass[acmsmall]{acmart}

\usepackage{booktabs}
\usepackage{rotating}
\usepackage{adjustbox}
\usepackage{diagbox}
\usepackage{multirow}
\usepackage{caption}
\usepackage[flushleft]{threeparttable} 

\usepackage{graphicx}
\usepackage{epstopdf}
\graphicspath{{figures/}}
\usepackage{subfigure}
\usepackage{stfloats}

\usepackage{amsmath}
\usepackage{amsmath}
\usepackage{bm}
\usepackage{algorithm}
\usepackage{algorithmic}
\usepackage{listings}
\usepackage{array}

\def\BibTeX{{\rm B\kern-.05em{\sc i\kern-.025em b}\kern-.08emT\kern-.1667em\lower.7ex\hbox{E}\kern-.125emX}}

\begin{document}

%
\title{Multiple Face Analyses through Adversarial Learning}

%
\author{Shangfei Wang}\authornote{Dr. Shangfei Wang is the corresponding author.}
\email{sfwang@ustc.edu.cn}
\affiliation{%
  \institution{University of Science and Technology of China}
  \streetaddress{443 HuangShan Rd}
  \city{Hefei Shi}
  \state{Anhui Sheng}
  \country{China}
  \postcode{230027}
}

\author{Shi Yin}
\affiliation{%
	\institution{University of Science and Technology of China}
	\streetaddress{443 HuangShan Rd}
	\city{Hefei Shi}
	\state{Anhui Sheng}
	\country{China}
	\postcode{230027}
}
\email{davidyin@mail.ustc.edu.cn}

\author{Longfei Hao}
\affiliation{%
  \institution{University of Science and Technology of China}
  \streetaddress{443 HuangShan Rd}
  \city{Hefei Shi}
  \state{Anhui Sheng}
  \country{China}
  \postcode{230027}
}
\email{hlf101@mail.ustc.edu.cn}

\author{Guang Liang}
\affiliation{%
  \institution{University of Science and Technology of China}
  \streetaddress{443 HuangShan Rd}
  \city{Hefei Shi}
  \state{Anhui Sheng}
  \country{China}
  \postcode{230027}
}
\email{xshmlgy@mail.ustc.edu.cn}

%

%
\begin{abstract}
    This inherent relations among multiple face analysis tasks, such as landmark detection, head pose estimation, gender recognition and face attribute estimation are crucial to boost the performance of each task, but have not been thoroughly explored since typically these multiple face analysis tasks are handled as separate tasks. In this paper, we propose a novel deep multi-task adversarial learning method to localize facial landmark, estimate head pose and recognize gender jointly or estimate multiple face attributes simultaneously through exploring their dependencies from both image representation-level and label-level. Specifically, the proposed method consists of a deep recognition network $\mathcal{R}$ and a discriminator $\mathcal{D}$. The deep recognition network is used to learn the shared middle-level image representation and conducts multiple face analysis tasks simultaneously. Through multi-task learning mechanism, the recognition network explores the dependencies among multiple face analysis tasks, such as facial landmark localization, head pose estimation, gender recognition and face attribute estimation from image representation-level. The discriminator is introduced to enforce the distribution of the multiple face analysis tasks to converge to that inherent in the ground-truth labels. During training, the recognizer tries to confuse the discriminator, while the discriminator competes with the recognizer through distinguishing the predicted label combination from the ground-truth one. Though adversarial learning, we explore the dependencies among multiple face analysis tasks from label-level. Experimental results on four benchmark databases, i.e., the AFLW database, the Multi-PIE database, the CelebA database and the LFWA database, demonstrate the effectiveness of the proposed method for multiple face analyses.
\end{abstract}

%
%

%

%
\maketitle

\section{Introduction}\label{sec:introduction}
\begin{figure}[!h]
  \centering
  \subfigure[]{\includegraphics[width=0.19\linewidth]{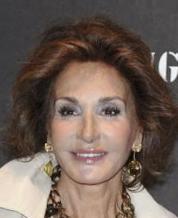}}
  \subfigure[]{\includegraphics[width=0.19\linewidth]{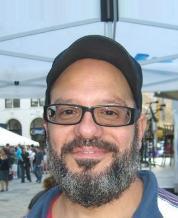}} \\
  \subfigure[]{\includegraphics[width=0.19\linewidth]{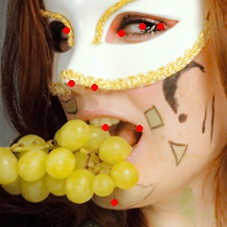}}
  \subfigure[]{\includegraphics[width=0.19\linewidth]{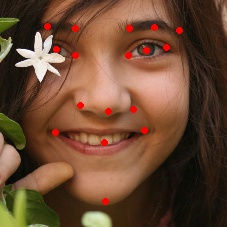}}
  \subfigure[]{\includegraphics[width=0.19\linewidth]{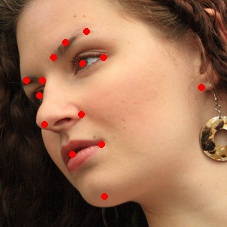}}
  \caption{The dependencies among multiple face analysis tasks.}\label{fig:landmark-occ}
\end{figure}

Face analyses have attracted increasing attention in recent years due to their wide applications in human computer interaction. Face analyses include several tasks, such as facial landmark detection, head pose estimation, face identification, facial expression classification, gender recognition and multiple face attribute estimation. These tasks are related to each other. For example, as shown in Fig.~\ref{fig:landmark-occ}, a person who wears necklace and earrings is more likely to be a female, and is less likely to be a male; and a person with sideburns and goatee is more likely to be a male, and is less likely to be a female; the locations of landmark are affected by head poses; facial expression variations obviously influence the location of landmarks. Such inherent connections among facial landmarks, head poses and expressions or multiple face attributes can be leveraged for multiple face analysis tasks, but have not been thoroughly explored yet, since typically face analysis tasks are handled separately.

Only very recently, a few works have turned to solve several face analysis tasks jointly. Zhang~\emph{et al.}~\cite{zhang2014facial} and  Ranjan~\emph{et al.}~\cite{ranjan2017hyperface} modeled dependencies among several face analysis tasks from the learned representation-level. Zhang~\emph{et al.}~\cite{zhang2014panda} proposed a multi-task convolutional neural network (CNN) consisting of shared features for heterogeneous face attributes. But they failed to consider task dependencies inherent in label-level. Zhu and  Ramanan~\cite{zhu2012face} considered the task dependencies from the label-level, but ignored their dependencies in  facial appearance. Instead of  jointly learning multiple face analysis tasks in a parallel way, like the above work,  Wu~\emph{et al.}~\cite{wu2017simultaneous} and  Honari~\emph{et al.}~\cite{honari2018improving} tried to leverage task dependencies in a serial manner, and mainly captured multi-task dependencies in representation-level.

To the best of our knowledge, although both representation-level dependencies and label-level dependencies are critical for multiple face analysis tasks, little work addresses them simultaneously till now. Therefore, in this paper, we propose a deep multi-task adversarial learning method for multiple face analysis tasks through exploring their dependencies from both representation-level and label-level. Specifically, we construct a deep network as a multi-task recognizer to explore connections among multiple tasks through representation-level. Then, we introduce adversarial loss to enforce the joint distribution of the labels predicted by the multi-task recognizer converge to that inherent in the ground-truth labels, and thus leverage multi-task dependencies from the label-level. Experimental results on four benchmark databases demonstrate that the proposed method successfully leverages task dependencies inherent in both representation and target label and thus achieves state of the art performance on multiple face analysis tasks.

The rest of this paper is organized in the following manner. Section~\ref{sec:related-work} gives an overview of the related work on multiple face analyses. Section~\ref{sec:problem} briefly gives the problem statement for our method. Section~\ref{sec:method} elaborates on the proposed method for multiple face analyses. Section~\ref{sec:experiment} presents the experimental results and analyses on four databases, and makes the comparison to related works. Section~\ref{sec:conclusion} concludes our work.

\section{Related work}\label{sec:related-work}
In this section, we summarize and analyze recent multiple face analysis works. We divide these works into two categories: i.e. facial landmark related multiple face analyses and multiple face attribute estimation. Furthermore, we discuss recent work on adversarial multi-task learning.

\subsection{Facial landmark related multiple face analyses}
Facial landmark detection is a prerequisite for many face analysis tasks, such as head pose estimation, face recognition, facial expression recognition and gender recognition. A comprehensive survey of facial landmark detection can be found in Wu and Ji's work~\cite{wu2017facial}. In this section, we discuss several recent works of landmark related multiple face analyses.

Zhang~\emph{et al.}~\cite{zhang2014facial} proposed a task-constrained deep convolutional network (TCDCN) to jointly optimize facial landmark detection with a set of related tasks, such as pose estimation, gender recognition, glasses detection, and smiling classification. They further systematically demonstrated that  the representations learned from related tasks facilitate the learning of facial landmark detector.  Instead of pooling from  the same feature space for all tasks as Zhang~\emph{et al.}~\cite{zhang2014facial} did, Ranjan~\emph{et al.}~\cite{ranjan2017hyperface} strategically designed the network architecture to exploit both low-level and high-level features of the network. They proposed HyperFace, a deep multi-task learning framework for simultaneous face detection, landmarks localization, pose estimation and gender recognition.

Unlike Zhang~\emph{et al.}~\cite{zhang2014facial}'s and Ranjan~\emph{et al.}~\cite{ranjan2017hyperface}'s works, which explored the inherent dependencies among multiple face analysis tasks from the learned representation-level, Zhu and  Ramanan~\cite{zhu2012face} considered the dependencies from the label-level, i.e. the topological changes due to related factors. They proposed a method for face detection, pose estimation, and landmark localization (FPLL) simultaneously. Specifically, they proposed  a mixtures of trees with a shared pool of parts. Every facial landmark is modeled as a part, and the topological changes due to viewpoint are captured by the global mixtures.

Instead of jointly learning multiple face analysis tasks in a parallel way, like the above works,  Wu~\emph{et al.}~\cite{wu2017simultaneous} proposed an iterative cascade method to simultaneously perform facial landmark detection, pose and deformation estimation. Their method iteratively updated the facial landmark locations, facial occlusion, head pose and facial deformation until convergence. Although the iterative cascade procedure can capture connections among multiple face analysis tasks at representation-level, the errors caused in the previous iteration may be propagated to the next iteration. Therefore, we prefer to jointly learning multiple face analysis tasks in a parallel way.

Other than exploring task dependencies in supervised learning scenarios, Honari~\emph{et al.}~\cite{honari2018improving} leveraged task dependencies to improve landmark localization in semi-supervised learning scenarios.  They proposed a framework of sequential multitasking learning for landmark localization and related face analysis tasks, such as expression recognition. Specifically, their proposed method first detected landmarks, and then the detected landmarks are  the input of the related face analysis tasks, which are  acted as an auxiliary signal to guide the landmark localization on unlabeled data. Although their proposed sequential multitasking learning framework successfully explores related face analysis tasks to boost facial landmark detection under partially labeled data, the dependencies among tasks are mainly exploited in the learned representation-level, not in the label-level. Furthermore, the errors caused by the first stage could be propagated to the next stage, and vice versa.

To the best of our knowledge, few works leverage inherent dependencies among landmark-related multiple face analysis tasks from both representation-level and label-level. Therefore, we propose a deep multi-task adversarial learning method for facial landmark detection enhanced by multiple face analysis tasks through exploring their dependencies from both representation-level and label-level. Specifically, we first construct a deep network as a multi-task recognizer $\mathcal{R}$ to jointly detect  facial landmarks, estimate landmark visibility, recognize head pose and classify gender. Through multi-task learning, the designed deep network can explore connections among multiple tasks through representation-level. Then, we introduce a discriminator $\mathcal{D}$ to distinguish the ground-truth label combination from the output of the recognizer $\mathcal{R}$. During training, $\mathcal{R}$ maximums the probability of mistake made by $\mathcal{D}$, while $\mathcal{D}$ does the opposite. Through such adversarial learning, the proposed method enforces the joint distribution of the labels predicted by $\mathcal{R}$ converge to that inherent in the ground-truth labels, and thus leverages multi-task dependencies from the label-level.

\subsection{Multiple face attribute estimation}
Face attribute estimation has attracted increasing attentions, since face attributes are middle-level abstraction between the low-level facial features and the high-level labels. FaceTacker~\cite{kumar2008facetracer} used a combination of support vector machines and Adaboost to select the optimal features for each attribute, and train each attribute classifier separately. It ignores the relations among multiple face attributes, which can be leveraged to boost the performance of multiple face attribute estimation. Zhang~\emph{et al.}~\cite{zhang2014panda} proposed Pose Alignment Networks for Deep Attribute modeling (PANDA) to obtain a pose-normalized deep representation for multiple face attribute estimation.
Liu~\emph{et al.}~\cite{liu2015deep} believed  that face localization can improve the performance of multiple face attribute estimation, and thus cascaded face localization networks (LNets) and the attribute network (ANet). Zhong~\emph{et al.}~\cite{zhong2016face} combined several off-the-shelf convolutional neural networks (i.e., CTS-CNN), which are trained for face recognition to estimate multiple face attributes simultaneously. These above works explore the  shared representations  for multiple face attribute estimation, but ignore the dependencies among multiple face attributes from the label-level.

Han~\emph{et al.}~\cite{han2018heterogeneous} tried to model both representation-level and label-level dependencies. They proposed  a CNN to capture representation-level dependencies through the shared low-level features for all attributes and task specific high-level features for heterogeneous attributes. They further proposed constraints according to prior knowledge to capture the fixed label-level dependencies. Instead of using constrains to model fixed dependencies,  Hand~\emph{et al.}~\cite{hand2017attributes} proposed a multi-task CNN (MCNN-AUX) to learn the label-level dependencies through an auxiliary network stacked on the top. Cao~\emph{et al.}~\cite{cao2018partially} considered the identity information and attribute relationships jointly. They proposed a partially Shared Multi-task CNN (PS-MCNN) to learn the task specific and shared features, and then utilized the identity information to improve the performance of face attribute estimation (PS-MCNN-LC). Although the above three works can explore dependencies from both representation-level and label-level for multiple face attribute estimation, the captured label-level dependencies are either fixed or represented by fixed form through the structure and parameters of a network.

To address the above issues, the proposed work employs an adversarial strategy to  capture label distributions directly without the assumption of the distribution form. Specifically, we first construct a deep multi-task network  $\mathcal{R}$ to estimate multiple face attributes simultaneously. Then, we introduce a discriminator $\mathcal{D}$ to distinguish the ground-truth label combination from the output of the recognizer $\mathcal{R}$. Through adversarial learning, the proposed method leverages multi-task dependencies from both label-level and representation-level to facilitate multiple face attribute estimation.

\subsection{Adversarial multi-task learning}
Recent years have seen a few works incorporating adversary learning with multi-task learning. For example, Bai~\emph{et al.}~\cite{bai2018sod} introduced  a generator to up-sample small blurred images into fine-scale ones for more accurate detection, and a discriminator describes each super-resolution image patch with multiple scores. Liu~\emph{et al.}~\cite{liu2017adversarial} proposed to alleviate the shared and private latent feature spaces from interfering with each other by using adversarial training and orthogonality constraints. The adversarial training is used to construct common and task-invariant shared latent spaces, while the orthogonality constraint is used to eliminate redundant features from the private and shared spaces. Liu~\emph{et al.}~\cite{liu2018multi} proposed an encoder to extract a disentangled feature representation for the factors of interest, and the discriminators to classify each of the factors as individual tasks. The encoder and the discriminators are trained cooperatively on factors of interest, but in an adversarial way on factors of distraction. All above works leverage adversarial learning for better input data or representations for multi-task learning, but ignore the dependencies among target labels. We are the first to explore dependencies among multiple tasks from both representation and label-level through adversarial mechanism.

\section{Problem Statement}\label{sec:problem}
Let $\bm{T} = \{\bm{x}, \bm{y} \}^N$ denotes $N$ training samples, where $\bm{x}$ represents the facial image, $\bm{y} = \{\bm{t}_1, \bm{t}_2, \dots, \bm{t}_n \}$ represents the ground-truth labels, such as facial landmark locations, visibility of each landmark, head pose angle and gender information or multiple face attributes. The purpose of the paper is to learn a multi-task recognizer  $\mathcal{R}: \bm{x}\rightarrow\bm{y}$ through optimizing the following formula:

\begin{equation}\label{eq:obj}
    \min_{\Theta_{\mathcal{R}}} \alpha_1 * \mathcal{L}_{s}(\mathcal{R}(\bm{x}; \Theta ), \bm{y}) + \alpha_2 * \mathcal{L}_{d}(P_{\bm{y}}, P_{\bm{y'}}) \, ,
\end{equation}

where $\mathcal{L}s$ is the supervised loss of multiple tasks, $\Theta_{\mathcal{R}}$ are parameters of multi-task recognizer $\mathcal{R}$, $\bm{y'} = \mathcal{R}(\bm{x})$, $P_{\bm{y}}$ and $P_{\bm{y'}}$ are the distribution of the ground-truth label and the distribution of the predicted labels from $\mathcal{R}$, respectively, $\mathcal{L}_{d}$ is the distance between two distributions. The first term minimizes the recognition errors of multi-tasks that sharing common representations, and the second term closes the joint distribution of the predicted label combination to the ground-truth label combination. $\alpha_1$ and $\alpha_2$ balance these two terms. Therefore, the proposed method can successfully explore connections among multiple face analysis tasks through both representation-level and label-level.

\section{Proposed Method}\label{sec:method}
\begin{figure}[!t]
	\centering
	\includegraphics[width=0.96\linewidth]{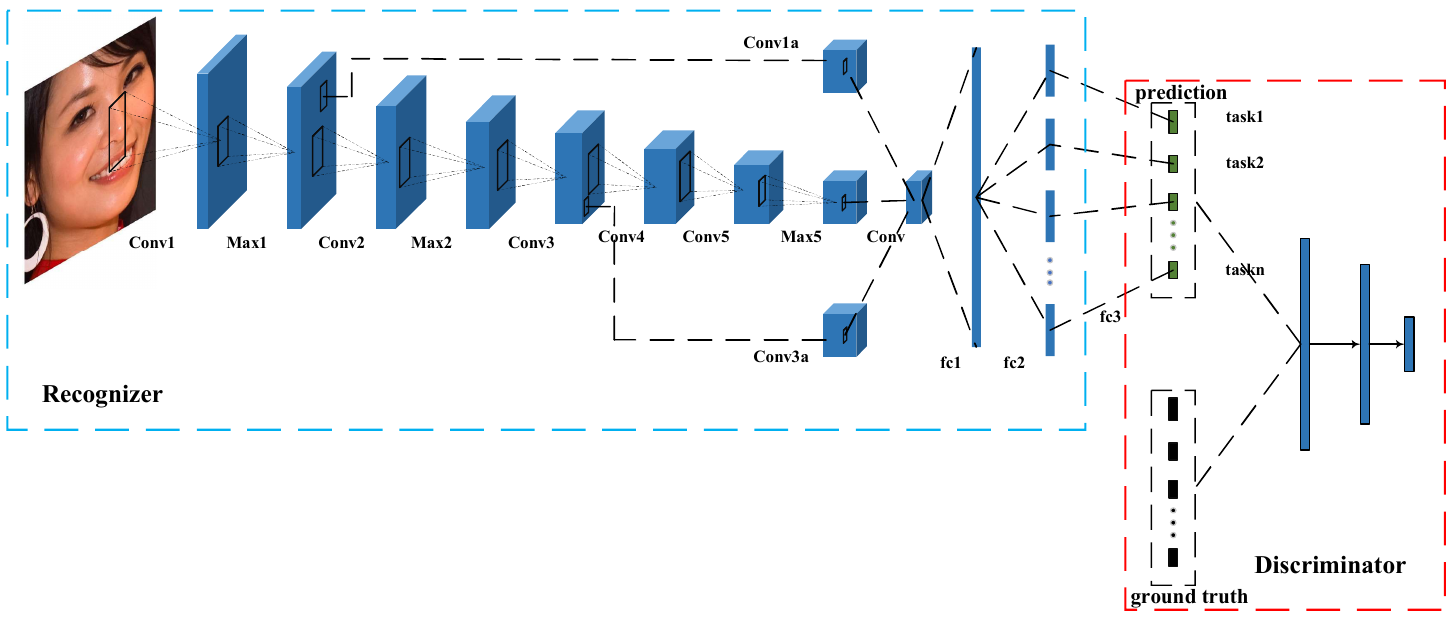}
	\caption{The framework of proposed model.}
	\label{fig:framework}
\end{figure}

The framework of the proposed deep multi-task adversarial learning method is shown in Fig.~\ref{fig:framework}. It consists of a deep multi-task recognizer $\mathcal{R}$ and a discriminator $\mathcal{D}$. The goal of $\mathcal{R}$ is to learn shared image representation and predict landmarks, visibility, pose and gender simultaneously or multiple face attributes simultaneously. $\mathcal{D}$ is to distinguish the ground-truth label combination from the label combination predicted by $\mathcal{R}$.
With the supervisory information of the ground-truth label combinations, the recognizer $\mathcal{R}$ can successfully capture the connections among multiple face analysis tasks by sharing feature representations. Through the competition between $\mathcal{R}$ and $\mathcal{D}$, the distribution of the predicted label combination could converge to the label distribution of the ground-truth. Thus, the proposed method can model the dependencies among landmark, visibility, pose, gender and the dependencies among multiple face attributes.

Through adversarial learning, we can minimize the distance of two distributions, i.e., the second term of Equation~\ref{eq:obj}, but do not need to model $P_{\bm{y}}$ and $P_{\bm{y'}}$ directly, which are complex and error prone processes. We replace $\mathcal{L}_{d}(P_{\bm{y}}, P_{\bm{y'}})$ as the following adversarial loss:

\begin{equation}\label{adv}
\min_{\mathcal{R}}\max_{\mathcal{D}}\mathcal{L}_{adv}=\mathbb{E}_{\bm{y}}[\log \mathcal{D}(\bm{y})]+\mathbb{E}_{\bm{\hat{y}}}[\log (1-\mathcal{D}(\bm{\hat{y}}))] \, ,
\end{equation}

where $\bm{\hat{y}}=\mathcal{R}(\bm{x})$ is the predicted label combination of facial image $\bm{x}$ that is regard as ``fake'', $\bm{y}$ is the ground-truth label combination regarded as ``real''. It's hard to optimize the above problem directly. We seek individual objective for $\mathcal{R}$ and $\mathcal{D}$ as Goodfellow~\emph{et al.}~\cite{goodfellow2014generative} did and utilize an alternate training procedure as described in the following sections.

\subsection{Recognizer}
One objective of recognizer $\mathcal{R}$ is to minimize $\mathcal{L}_{adv}$ in Equation~\ref{adv}. It means recognizer $\mathcal{R}$ tries to `fool' discriminator $\mathcal{D}$ and let it classify the predicted label combination $\bm{\hat{y}}$ as ``real''. Therefore, the adversarial objective for recognizer $\mathcal{R}$ is as follows:

\begin{equation}
\mathcal{L}_{adv}^{\mathcal{R}} = -\log\mathcal{D}(\bm{\hat{y}}) \, .
\end{equation}

We formulate multiple face analysis tasks as a multi-task learning problem. It indicates landmark detection, visibility recognition, head pose estimation and  gender recognition or multiple face attribute estimation. Therefore, we construct a multi-task recognizer $\mathcal{R}$ to learn the sharing representation and explore dependencies among these multiple tasks. The supervised loss $\mathcal{L}_{s}$ for multi-task recognizer $\mathcal{R}$ contains the following losses:

\vspace{5pt}
\textbf{Landmark Detection:}  The supervised loss for landmark detector is described as Equation~\ref{eq:loss-landmark}.

\begin{equation}\label{eq:loss-landmark}
\mathcal{L}_{s}^{L} = \frac{1}{2m} \sum_{i=1}^{m} v_i ((\hat{x_i} - x_i)^2 + (\hat{y_i}-y_i)^2) \, ,
\end{equation}

where $(x_i, y_i)$ is the location of $i^{th}$ landmark, $(\hat{x_i}, \hat{y_i})$ is corresponding estimation, $m$ is the total number of landmark points in one image. The visibility factor $v_i$ is $1$ if the $i^{th}$ landmark is visible, otherwise is $0$, which implies that the ground-truth location for $i^{th}$ landmark is not provided.

\vspace{5pt}
\textbf{Visibility Recognition:} We learn the visibility recognizer to predict the visibility of all landmarks. $\bm{v}$ is a multiple binary-value label vector. Hence, the supervised loss for visibility recognizer is shown as in Equation~\ref{eq:loss-visibility}:

\begin{equation}\label{eq:loss-visibility}
\mathcal{L}_{s}^{V} = -\frac{1}{m} \sum_{i=1}^{m} (v_i \log \hat{v_i} + (1-v_i) \log (1-\hat{v_i})) \, ,
\end{equation}

where $\hat{v_i}$ and $v_i$ are the predicted visibility and the corresponding the ground-truth visibility of $i^{th}$ landmark, respectively.

\vspace{5pt}
\textbf{Pose Estimation:} Since the pose information provided by database constructors are either continuous or discrete, the form of loss for pose estimator varies by databases. For continuous pose information (i.e., roll, pitch and yaw), the L2 loss function is used:

\begin{equation}\label{eq:loss-pose1}
   \mathcal{L}_{s}^{P} = \frac{1}{3}\left[(\hat{p_1}-p_1)^2 + (\hat{p_2}-p_2)^2 + (\hat{p_3}-p_3)^2\right] \, ,
\end{equation}

where $(p_1, p_2, p_3)$ are the ground-truth roll, pitch and yaw respectively, and $(\hat{p_1}, \hat{p_2}, \hat{p_3})$ are the estimated pose angles. For discrete pose information, we view the pose estimation as a multi-class classification problem and the cross-entropy loss is used.

\begin{equation}\label{eq:loss-pose2}
    \mathcal{L}_{s}^{P} = -\sum_{i=1}^{K} p_i \log (\hat{p_i}) \, ,
\end{equation}
where $(p_1, p_2, \dots, p_K)$ is the one-hot code of the ground-truth pose angle and $(\hat{p_1}, \hat{p_2}, \dots, \hat{p_K})$ is the one-hot code of corresponding estimated angle. $K$ is the number of angles.

\vspace{5pt}
\textbf{Gender Classification:} Gender classification is a binary classification problem. Hence, the supervised loss for gender classifier is as shown in Equation~\ref{eq:loss-gender}.

\begin{equation}\label{eq:loss-gender}
    \mathcal{L}_{s}^{G} = - \left[g \log(\hat{g}) + (1 - g) \log (1-\hat{g})\right] \, ,
\end{equation}

where $\hat{g}$ and $g$ are the predicted gender and the corresponding the ground-truth gender, respectively.

\vspace{5pt}
\textbf{Face Attribute Estimation:} The face attributes are all binary. Therefore, the supervised loss for multiple face attribute estimator is shown as Equation~\ref{eq:loss-attr}:

\begin{equation}\label{eq:loss-attr}
	\mathcal{L}_{s}^{A} = - \frac{1}{n} \sum_{i=1}^{n} a_i \log \hat{a_i} + (1-a_i) \log (1-\hat{a}_i) \, ,
\end{equation}
where $\hat{a_i}$ and $a_i$ are the $i$th predicted face attribute and the corresponding the ground-truth attribute, respectively. $n$ is the number of attributes.

Finally, the full supervised loss $\mathcal{L}_{s}$ can be written as follows:
\begin{equation}\label{eq:loss-total1}
\mathcal{L}_{s}= \alpha_{L}*\mathcal{L}_{s}^{L}+\alpha_{V} *\mathcal{L}_{s}^{V}+\alpha_{P} *\mathcal{L}_{s}^{P}+\alpha_{G} *\mathcal{L}_{s}^{G} \, ,
\end{equation}

or

\begin{equation}\label{eq:loss-total2}
	\mathcal{L}_{s} = \alpha_{attr} * \mathcal{L}_{s}^{A} \, .
\end{equation}

We combine the supervised loss $\mathcal{L}_{s}$ and the adversarial loss $\mathcal{L}_{adv}^{\mathcal{R}}$ as the full objective of multi-task recognizer $\mathcal{R}$, shown as Equation~\ref{eq:loss-total}:

\begin{equation}\label{eq:loss-total}
\mathcal{L}^{R}= \mathcal{L}_{s}+\alpha_{A}*\mathcal{L}_{adv}^{\mathcal{R}} \, ,
\end{equation}

where $\alpha_{L}$, $\alpha_{V}$, $\alpha_{P}$, $\alpha_{G}$, $\alpha_{attr}$ and $\alpha_{A}$ are weight coefficients of supervised losses of corresponding subtasks and adversarial loss, respectively.

\subsection{Discriminator}
As shown in the right part of Fig.~\ref{fig:framework}, we construct a discriminator $\mathcal{D}$. The purpose of the discriminator is to classify the ground-truth label combination as ``real'' and the predicted label combination as ``fake''. Therefore, the adversarial loss for $\mathcal{D}$ is shown as Equation~\ref{eq:loss-d}:

\begin{equation}\label{eq:loss-d}
\mathcal{L}^{\mathcal{D}} = -[\log \mathcal{D}(\bm{y}) + \log (1-\mathcal{D}(\bm{\hat{y}}))]
\end{equation}

The multi-task recognizer $\mathcal{R}$ and the discriminator $\mathcal{D}$ are updated with an alternate procedure: fix $\mathcal{R}$, update $\mathcal{D}$ according to Equation~\ref{eq:loss-d}, and then fix $\mathcal{D}$, update $\mathcal{R}$ according to Equation~\ref{eq:loss-total}. This process repeats until convergence. During training, the minibatch stochastic gradient descent is adopted. The detailed training procedure is described in Algorithm~\ref{alg:gan}.

\begin{algorithm}[!htb]
	\caption{Training algorithm of the proposed multi-task adversary learning.}\label{alg:gan}
	\begin{algorithmic}[1]
		\item[\textbf{Input}] The training set $\bm{T}$, the batch size $s$, the number of training step $K$ and the hyper parameter $k$.
		\item[\textbf{Output}] The multi-task recognizer $\mathcal{R}$.
        \STATE Initialize the parameters $\Theta_{\mathcal{R}}$ and $\Theta_{\mathcal{D}}$ of $\mathcal{R}$ and $\mathcal{D}$, respectively.
		\FOR{$i= 1 \to K$}			
		\FOR{$j = 1 \to k$}
		\STATE Randomly sample mini-batch of $s$ facial images $\{\bm{x}\}^{s}_{i=1}$ from feature space and sample mini-batch of $s$ labels $\{\bm{y}\}^{s}_{i=1}$ from label space.
		\STATE Update the parameters of discriminator $\mathcal{D}$ by descending its gradient:
		\begin{equation*}
\nabla_{\Theta_{\mathcal{D}}}\left(-\frac{1}{s} \sum_{i=1}^{s} [\log \mathcal{D}(\bm{y}) + \log (1 - \mathcal{D}(\mathcal{R}(\bm{x})))]\right)
        \end{equation*}
		\ENDFOR
        \STATE Randomly sample a mini-batch of $s$ samples $\{\bm{x}, \bm{y}\}^{s}_{i=1}$ from training set $\bm{T}$
		\STATE Update multi-task recognizer $\mathcal{R}$ by descending its gradient according to Equation~\ref{eq:loss-total}.
		\ENDFOR
	\end{algorithmic}
\end{algorithm}

For the structure of discriminator $\mathcal{D}$, we adopt feed-forward network with two hidden layers. As for the multi-task recognizer $\mathcal{R}$, we adopt a deep convolution network. All the stride of polling layer is 2, and the size of fc1 and fc2 are 3072 and 512. (kernel\_size, stride) of convolution layers are: Conv1(11, 4), Conv1a(4, 4), Conv2(5, 1), Conv3(3, 1), Conv3a(2, 2), Conv4(3, 1), Conv5(3, 1). All convolution layers are followed by a Batch normalization layer \cite{ioffe2015batch} and ReLU activation unit.

\section{Experiments}\label{sec:experiment}
\subsection{Experimental conditions}
To the best of our knowledge, only the Annotated Facial Landmark in the Wild (AFLW) database~\cite{koestinger2011annotated} and the CMU Multi-PIE Face (Multi-PIE) database~\cite{gross2010multi} contain facial landmarks, corresponding visibility, head poses and gender information simultaneously. Therefore, we evaluate the proposed adversary multi-task learning approach for facial landmark related multiple face analyses on these two databases. Furthermore, the large-scale CelebFaces Attributes (CelebA) database~\cite{liu2015deep} and the Labeled Faces in the Wild (LFWA) database~\cite{huang2008labeled} are used to evaluate the proposed adversary multi-task learning approach for multiple face attribute estimation.

The AFLW~\cite{koestinger2011annotated} database contains 25, 993 faces in 21, 997 real-world images with full pose, expression, ethnicity, age and gender variations. It provides annotations for 21 landmark points per face, along with the face bounding-box, face pose (i.e., roll, pitch and yaw) and gender.  Following the same sample selecting strategy as Ranjan~\emph{et al.}'s \cite{ranjan2017hyperface} work, we randomly select 1000 images for testing and then divide them into 3 subsets according to their absolute yaw angles.

The Multi-PIE database contains 337 subjects, captured under 15 views and 19 illuminations in four recording sessions for a total of more than 750,000 images. Among them, 6152 images are labeled with landmarks, whose number varies from 39 to 68, depending on their visibility. Following the same sample selecting strategy as Wu~\emph{et al.}'s \cite{wu2017simultaneous} work, we use the facial images from the first 150 subjects as training data and use the subjects with IDs between 151 and 200 as testing data.

The CelebA database is a large scale unconstrained face attribute database and contains more than 10, 000 identities, each of which has twenty images. There are more than 200, 000 images total.  The LFWA database has 13, 233 images of 5749 identities. Each image in the CelebA database and LFWA database is annotated with forty face attributes. Both databases are challenging for attribute estimation, with large variations in expressions, poses, races, illumination, background, etc. Following Liu~\emph{et al.}~\cite{liu2015deep}, we use the images of first 8,000 identities to train our model and the images of the last 1,000 identities to test our model on the CelebA database. For the LFWA database, we randomly split them into half and half for training and testing as Liu~\emph{et al.}~\cite{liu2015deep} did.

For the AFLW database, we calculate bounding boxes from face eclipse provided by the database constructor. For the Multi-PIE database, we detect bounding boxes through OpenCV~\cite{viola2001rapid}. After cropped, the facial images are resized to $227 \times 227 \times 3$. In order to obtain enough data and improve their generalization performance. We augment the training data through random shifting bounding box, resizing bounding box and jittering the color of facial images. For the CelebA and LFWA databases, the images are processed through resizing and color jittering. Model selection is adopted to select hyper parameters of the proposed methods.

On the AFLW database, we adopt the Normalized Mean Error (NME)~\cite{jourabloo2015pose} for landmark detection as HyperFace~\cite{ranjan2017hyperface} did. On the Multi-PIE database, to compare fairly with Wu~\emph{et al}~\cite{wu2017simultaneous}, the Mean Absolute Error (MAE)~\cite{lei2008face} is adopted for landmark detection. For pose estimation, the absolute degree error is adopted on the AFLW database and the accuracy is adopted on the Multi-PIE database. For visibility and gender, the accuracy is adopted. For the multiple face attribute estimation, the accuracy is adopted.

To validate the effectiveness of the proposed adversary multi-task network for landmark-related multiple face analyses, six methods are compared:
the method considering task dependencies from representation-level only ($\rm Ours_{no}$), which employs the first term of Equation~\ref{eq:obj}; the method considering landmark dependencies ($\rm Ours_{l}$), where only the landmark is fed into discriminator; the method considering joint distribution of landmark and visibility ($\rm Ours_{lv}$), where the landmark and visibility are fed into discriminator; the method considering joint distribution of landmark, visibility, and gender ($\rm Ours_{lvg}$), where the landmark, visibility and gender are fed into discriminator; the method considering joint distribution of landmark, visibility, and pose ($\rm Ours_{lvp}$), where the landmark, visibility, and pose are fed into discriminator; and the proposed method considering joint distribution of landmark, visibility, gender and pose ($\rm Ours_{all}$), where the landmark, visibility, gender and pose are all fed into discriminator. For multiple face attribute estimation on the CelebA and LFWA databases, we compare $\rm Ours_{no}$ and $\rm Ours_{gan}$.

\subsection{Experimental results and analyses of facial landmark related multiple face analyses}
\begin{figure}[!htb]
  \centering
  \subfigure[Pose Estimation Error for Roll]{
    \includegraphics[width=0.32\linewidth]{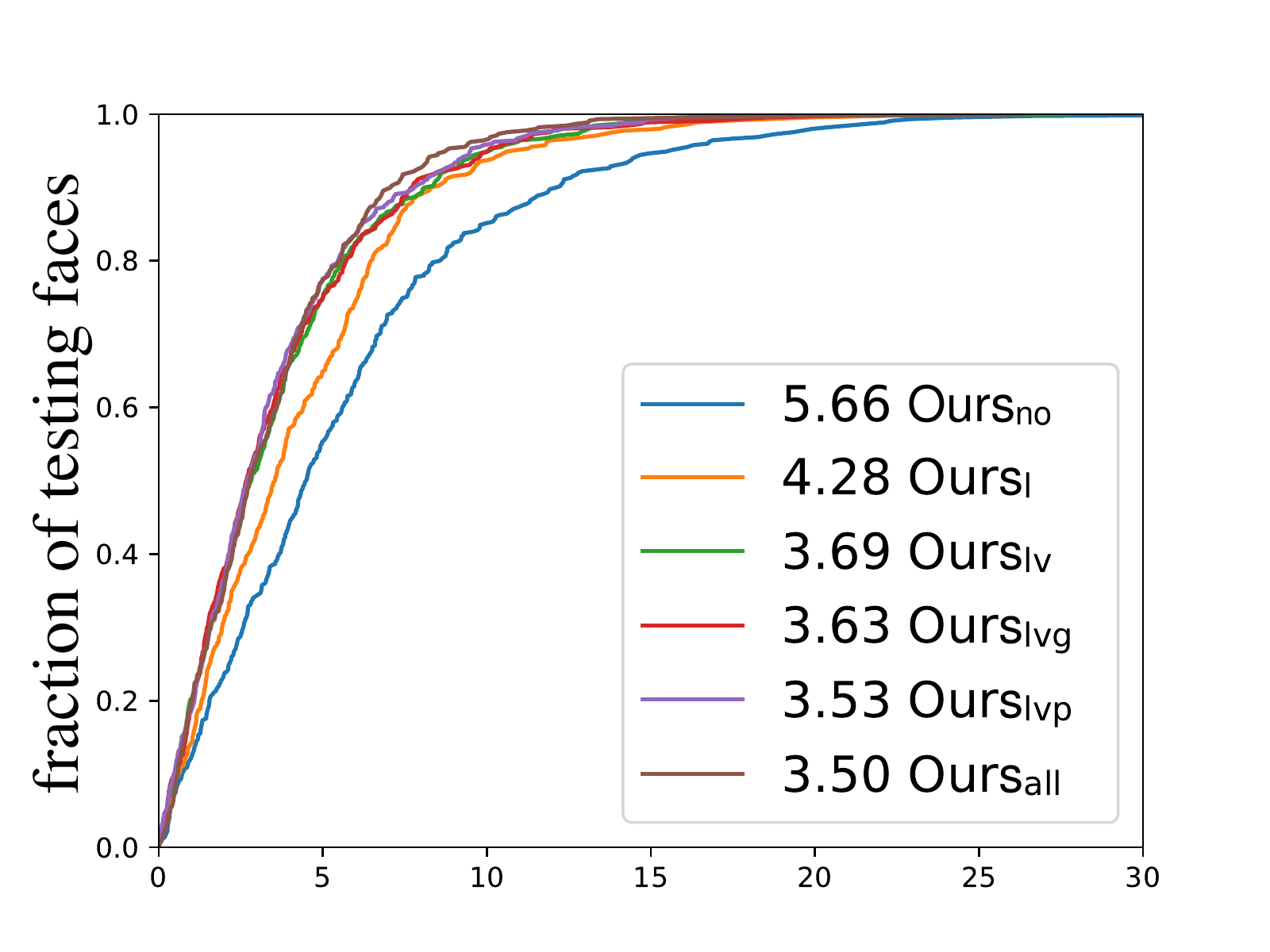}}
  \subfigure[Pose Estimation Error for Pitch]{
    \includegraphics[width=0.32\linewidth]{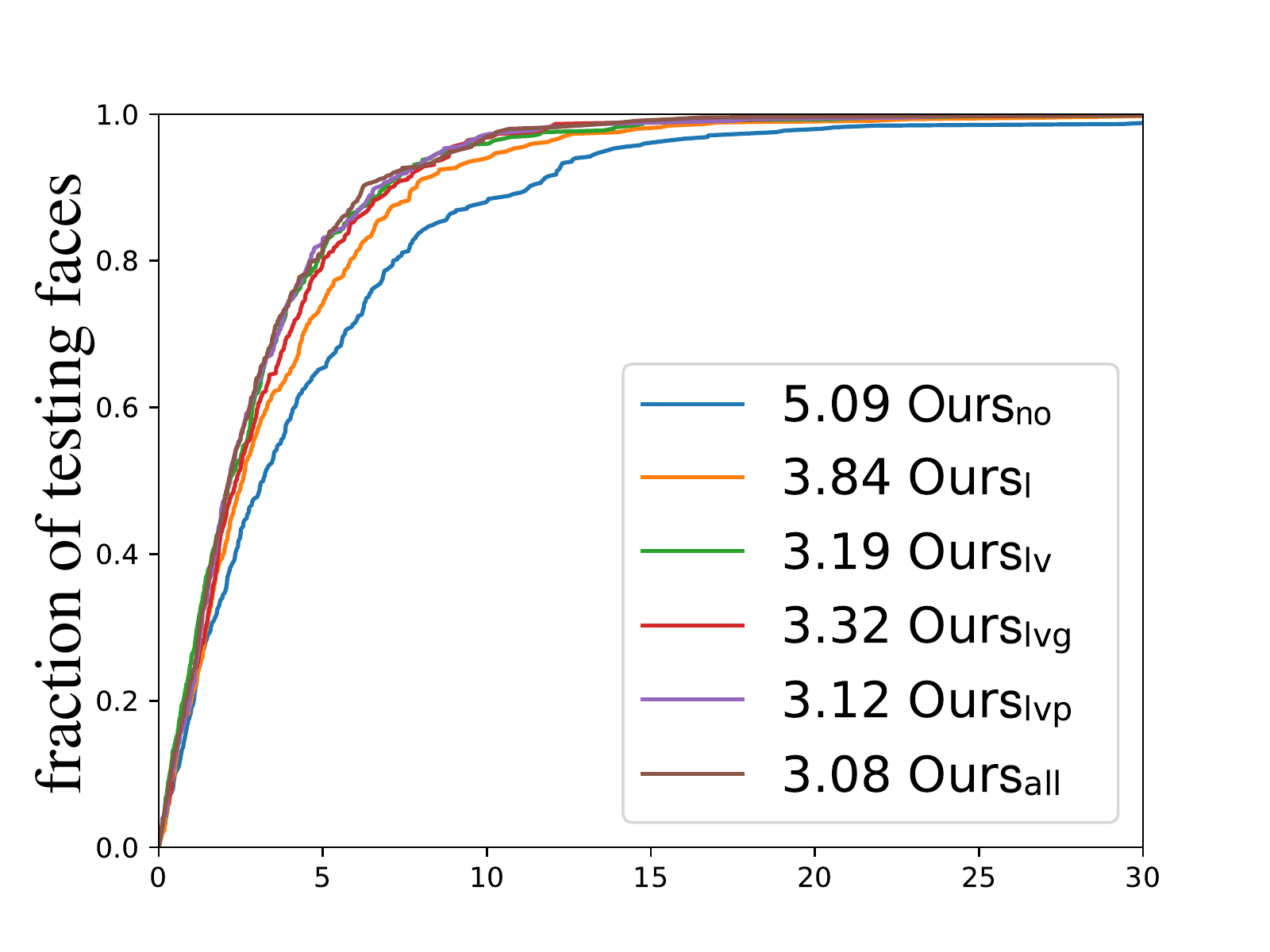}}
  \subfigure[Pose Estimation Error for Yaw]{
    \includegraphics[width=0.32\linewidth]{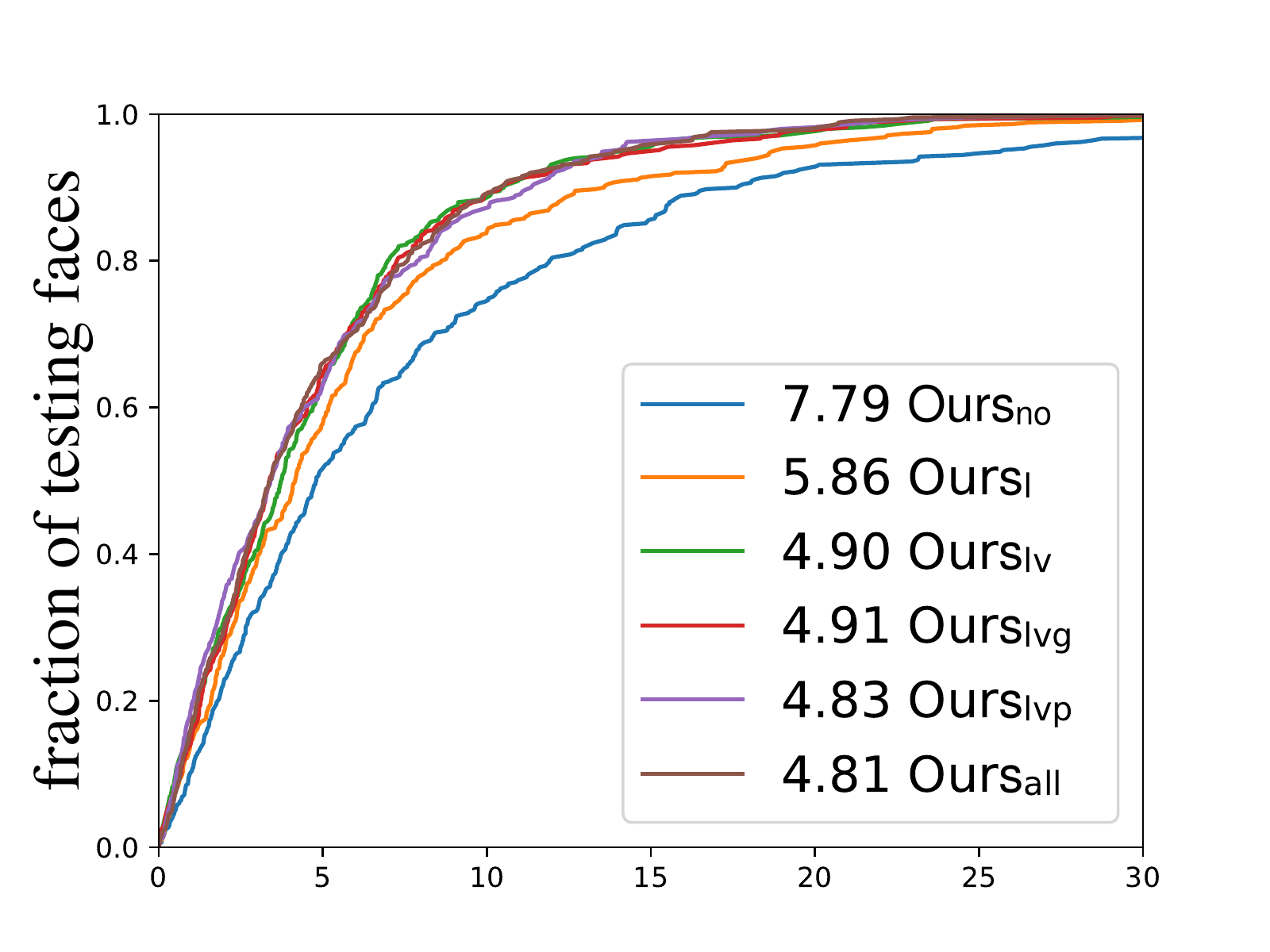}}
  \caption{Cumulative error curves of pose estimation on the AFLW database.}
  \label{fig:degree-err}
\end{figure}

Experimental results of landmark detection, visibility recognition and gender recognition on the AFLW and Multi-PIE databases are shown in Table~\ref{tab:res-alfw} and Table~\ref{tab:res-multipie}, respectively. Since the AFLW database provides 3 continuous poses, cumulative error curve is adopted for more detailed performance, as shown in Fig.~\ref{fig:degree-err}.

\begin{table}[!htbp]
	\caption{Experimental results on the AFLW database (21pts).}\label{tab:res-alfw}
  \centering
	\begin{tabular}{lll ccc c}
		\toprule
		Methods & Tasks & Metrics & [0, 30] & [30, 60] & [60, 90] & mean \\
		\midrule
		\multirow{3}{*}{$\rm Ours_{no}$} & landmark & NME($\downarrow$) & 3.70 & 3.75 & 4.33 & 3.93 \\
		& visibility & Acc($\uparrow$) & 0.93 & 0.92 & 0.95 & 0.93 \\
		& gender & Acc($\uparrow$) & 0.87 & 0.91 & 0.91 & 0.90 \\
		\midrule
		\multirow{3}{*}{$\rm{Ours}_{l}$} & landmark & NME($\downarrow$) & 3.41 & 3.68 & 4.20 & 3.76 \\
		& visibility & Acc($\uparrow$) & 0.95 & 0.94 & 0.97 & 0.95 \\
		& gender & Acc($\uparrow$) & 0.95 & 0.97 & 0.96 & 0.96 \\
        \midrule
		\multirow{3}{*}{$\rm{Ours}_{lv}$}& landmark & NME($\downarrow$) & 3.43 & 3.83 & 3.80 & 3.68 \\
		& visibility & Acc($\uparrow$) & 0.97 & 0.97 & 0.98 & 0.97 \\
		& gender & Acc($\uparrow$) & 0.97 & 0.97 & 0.96 & 0.97 \\
        \midrule
		\multirow{3}{*}{$\rm{Ours}_{lvg}$}& landmark & NME($\downarrow$) & 3.37 & 3.55 & 3.70 & 3.54 \\
		& visibility & Acc($\uparrow$) & 0.97 & 0.97 & 0.97 & 0.97 \\
		& gender & Acc($\uparrow$) & 0.97 & 0.96 & 0.97 & 0.97 \\
        \midrule
		\multirow{3}{*}{$\rm{Ours}_{lvp}$}&  landmark & NME($\downarrow$)  & 3.32 & 3.41 & 3.78 & 3.50 \\
		& visibility & Acc($\uparrow$) & 0.97 & 0.96 & 0.97 & 0.97 \\
		& gender & Acc($\uparrow$) & 0.97 & 0.95 & 0.97 & 0.96 \\
        \midrule
		\multirow{3}{*}{$\rm{Ours}_{all}$}& landmark & NME($\downarrow$) & 3.19 & 3.28 & 3.49 & 3.32\\
		& visibility & Acc($\uparrow$) & 0.97 & 0.96 & 0.97 & 0.97 \\
		& gender & Acc($\uparrow$) & 0.99 & 0.99 & 0.97 & 0.98 \\
		\bottomrule
	\end{tabular}

    \begin{tablenotes}
    \small
    \item \textit{Note:} $\uparrow$ represents that the higher value indicates better performance and $\downarrow$ represents that the smaller value indicates better performance.
    \end{tablenotes}
\end{table}

\begin{table}[!htbp]
    \tabcolsep=2pt
	\caption{Experimental results on the Multi-PIE database.}
	\label{tab:res-multipie}
  \centering
	\begin{tabular}{lll ccccccccccccc c}
		\toprule
		Methods & Tasks & Metrics & -90 & -75 & -60 & -45 & -30 & -15 & 0 & 15 & 30 & 45 & 60 & 75 & 90 & mean \\
		\midrule
		\multirow{4}{*}{$\rm Ours_{no}$} & landmark & MAE($\downarrow$) & 3.49 & 3.13 & 3.30 & 3.57 & 3.23 & 2.99 & 2.86 & 3.09 & 2.91 & 3.53 & 3.01 & 3.21 & 3.61 & 3.10 \\
		& visibility & Acc($\uparrow$) & 0.91 & 0.96 & 0.91 & 0.97 & 0.97 & 0.99 & 0.99 & 1.00 & 1.00 & 0.99 & 0.92 & 0.98 & 1.00 & 0.98 \\
		& pose & Acc($\uparrow$) & 1.00 & 0.93 & 0.96 & 0.95 & 1.00 & 1.00 & 1.00 & 0.99 & 0.98 & 1.00 & 1.00 & 1.00 & 1.00 & 0.99 \\
		& gender & Acc($\uparrow$) & 0.69 & 0.81 & 0.85 & 0.90 & 0.97 & 0.97 & 0.96 & 0.96 & 0.90 & 0.84 & 0.81 & 0.75 & 0.69 & 0.92 \\
		\midrule
		\multirow{4}{*}{$\rm{Ours}_{l}$} & landmark & MAE($\downarrow$) & 3.15 & 2.86 & 2.92 & 3.45 & 3.10 & 2.88 & 2.77 & 2.95 & 2.82 & 3.44 & 2.64 & 3.15 & 3.40 & 2.96 \\
		& visibility & Acc($\uparrow$) & 0.83 & 0.68 & 0.71 & 0.76 & 0.98 & 0.99 & 1.00 & 1.00 & 1.00 & 0.99 & 0.78 & 0.79 & 0.79 & 0.93 \\
		& pose & Acc($\uparrow$) & 1.00 & 1.00 & 0.93 & 1.00 & 1.00 & 1.00 & 1.00 & 0.99 & 0.98 & 1.00 & 1.00 & 1.00 & 1.00 & 0.99 \\
		& gender & Acc($\uparrow$) & 0.88 & 0.91 & 0.93 & 0.90 & 0.97 & 0.96 & 0.99 & 0.95 & 0.93 & 0.81 & 0.81 & 0.81 & 0.69 & 0.94 \\
        \midrule
		\multirow{4}{*}{$\rm{Ours}_{lv}$} & landmark & MAE($\downarrow$) & 3.11 & 2.86 & 2.95 & 3.45 & 3.08 & 2.87 & 2.75 & 2.86 & 2.81 & 3.47 & 2.53 & 3.26 & 3.47 & 2.94 \\
		& visibility & Acc($\uparrow$) & 1.00 & 0.99 & 0.96 & 0.99 & 0.99 & 0.99 & 1.00 & 1.00 & 1.00 & 1.00 & 0.92 & 0.99 & 1.00 & 0.99 \\
		& pose & Acc($\uparrow$) & 1.00 & 1.00 & 0.91 & 1.00 & 1.00 & 1.00 & 1.00 & 1.00 & 0.98 & 1.00 & 1.00 & 1.00 & 1.00 & 0.99 \\
		& gender & Acc($\uparrow$) & 0.88 & 0.87 & 0.94 & 0.88 & 0.97 & 0.96 & 0.98 & 0.91 & 0.90 & 0.83 & 0.81 & 0.81 & 0.69 & 0.93 \\
        \midrule
		\multirow{4}{*}{$\rm{Ours}_{lvg}$} & landmark & MAE($\downarrow$) & 2.89 & 2.77 & 2.98 & 3.45 & 3.13 & 2.81 & 2.70 & 2.90 & 2.76 & 3.38 & 2.63 & 3.07 & 3.36 & 2.91 \\
        & visibility & Acc($\uparrow$) & 0.96 & 0.90 & 0.82 & 0.99 & 0.99 & 0.99 & 1.00 & 1.00 & 1.00 & 0.99 & 0.57 & 0.64 & 0.65 & 0.96 \\
		& pose & Acc($\uparrow$) & 1.00 & 0.98 & 0.91 & 1.00 & 1.00 & 1.00 & 1.00 & 0.99 & 0.98 & 1.00 & 1.00 & 1.00 & 1.00 & 0.99 \\		
        & gender & Acc($\uparrow$) & 0.81 & 0.87 & 0.93 & 0.87 & 0.98 & 0.97 & 0.98 & 0.96 & 0.93 & 0.90 & 0.81 & 0.81 & 0.75 & 0.94 \\		
        \midrule
		\multirow{4}{*}{$\rm{Ours}_{lvp}$} & landmark & MAE($\downarrow$) & 3.00 & 2.80 & 2.95 & 3.42 & 3.09 & 2.80 & 2.71 & 2.84 & 2.71 & 3.37 & 2.57 & 3.02 & 3.35 & 2.90 \\
		& visibility & Acc($\uparrow$) & 1.00 & 0.97 & 0.92 & 0.99 & 0.99 & 0.99 & 0.99 & 1.00 & 1.00 & 1.00 & 0.80 & 0.85 & 0.91 & 0.98 \\
		& pose & Acc($\uparrow$) & 1.00 & 1.00 & 0.94 & 1.00 & 1.00 & 1.00 & 1.00 & 0.99 & 0.98 & 1.00 & 1.00 & 1.00 & 1.00 & 0.99 \\
		& gender & Acc($\uparrow$) & 0.88 & 0.91 & 0.93 & 0.90 & 0.94 & 0.95 & 0.98 & 0.91 & 0.90 & 0.86 & 0.81 & 0.75 & 0.75 & 0.93 \\
        \midrule
		\multirow{4}{*}{$\rm{Ours}_{all}$} & landmark & MAE($\downarrow$) & 3.02 & 2.68 & 2.82 & 3.39 & 3.07 & 2.78 & 2.68 & 2.76 & 2.69 & 3.26 & 2.46 & 2.94 & 3.20 & 2.85 \\
		 & visibility & Acc($\uparrow$) & 1.00 & 0.98 & 0.96 & 0.98 & 0.98 & 0.99 & 1.00 & 1.00 & 1.00 & 1.00 & 0.70 & 0.70 & 0.72 & 0.98 \\
		& pose & Acc($\uparrow$) & 1.00 & 0.98 & 0.98 & 1.00 & 1.00 & 1.00 & 1.00 & 0.99 & 0.98 & 1.00 & 1.00 & 1.00 & 1.00 & 1.00 \\
		& gender & Acc($\uparrow$) & 0.88 & 0.91 & 0.94 & 0.88 & 0.97 & 0.97 & 0.99 & 0.95 & 0.87 & 0.83 & 0.81 & 0.75 & 0.69 & 0.94 \\
		\bottomrule
	\end{tabular}

    \begin{tablenotes}
    \small
    \item \textit{Note:} $\uparrow$ represents that the higher value indicates better performance and $\downarrow$ represents that the smaller value indicates better performance.
    \end{tablenotes}
\end{table}

From Table~\ref{tab:res-alfw}, Table~\ref{tab:res-multipie} and Fig.~\ref{fig:degree-err}, we observe the following:

First, the experimental results for near frontal faces are better than those for other poses for all methods. Specifically, the experimental results for [0, 30] yaw angle on the AFLW database and the experimental results  for 0 view on the Multi-PIE database are the best, with the lowest error and the highest accuracy in the most cases. It is reasonable since face analyses from facial images with extreme head pose are more changeling than those from near frontal views.

Second, the proposed method considering both shared representation and label-level connection significantly outperforms the proposed method only exploiting representation-level connection. Specifically, on the AFLW database, compared to the proposed method only exploiting representation-level constraint , $\rm Ours_{l}$ decreases the average NME(\%) of the landmark detection by 0.17, and increases the accuracy of visibility and gender recognition by 2\% and 6\%, respectively. On the Multi-PIE database, $Ours_{l}$ decreases the average MAE of the landmark detection by 0.14, and increases the accuracy of gender recognition by 2\%. It is expected since the method considering both shared representation and label-level constraint models the inherent dependencies among multiple face analysis tasks more faithfully and completely than the method only exploiting shared representation.

Third, more label-level constraints may achieve better performance. Specifically, $\rm Ours_{all}$ performs best among the six methods, with the lowest NME or MAE for landmark detection, and the highest accuracy for visibility, gender recognition and pose estimation on both databases. It indicates that capturing more task relations from label-level can boost the performance of multiple tasks better.

Fourth, different label combinations lead to different effects. For instance, compared to $\rm Ours_{lvg}$,  $\rm Ours_{lvp}$ achieves lower error for landmark detection on both databases. The possible reason is that there exist more close relations between landmark and head poses than landmark and genders.

In addition, the improvement of the proposed method on the Multi-PIE database is weaker than that on the AFLW database. The Muti-PIE database consists of 13 fixed views, without pitch and roll angles, while the AFLW database provides continuous  pitch, roll and yaw angles in the real world. Therefore, it is easier to detect landmark, estimate poses and recognize genders on the Multi-PIE database than the AFLW database. This results in less improvement space on the Multi-PIE database. Another possible reason is that the proposed method successfully captures the constraint among multiple face analysis tasks through both representation-level and label-level and captured constraint provides more benefits for face analyses from more challenging data, like facial images collected in the wild.

\begin{figure}[!htbp]
	\centering
	\subfigure[]{
        \includegraphics[width=0.45\linewidth,trim=0 30 0 0]{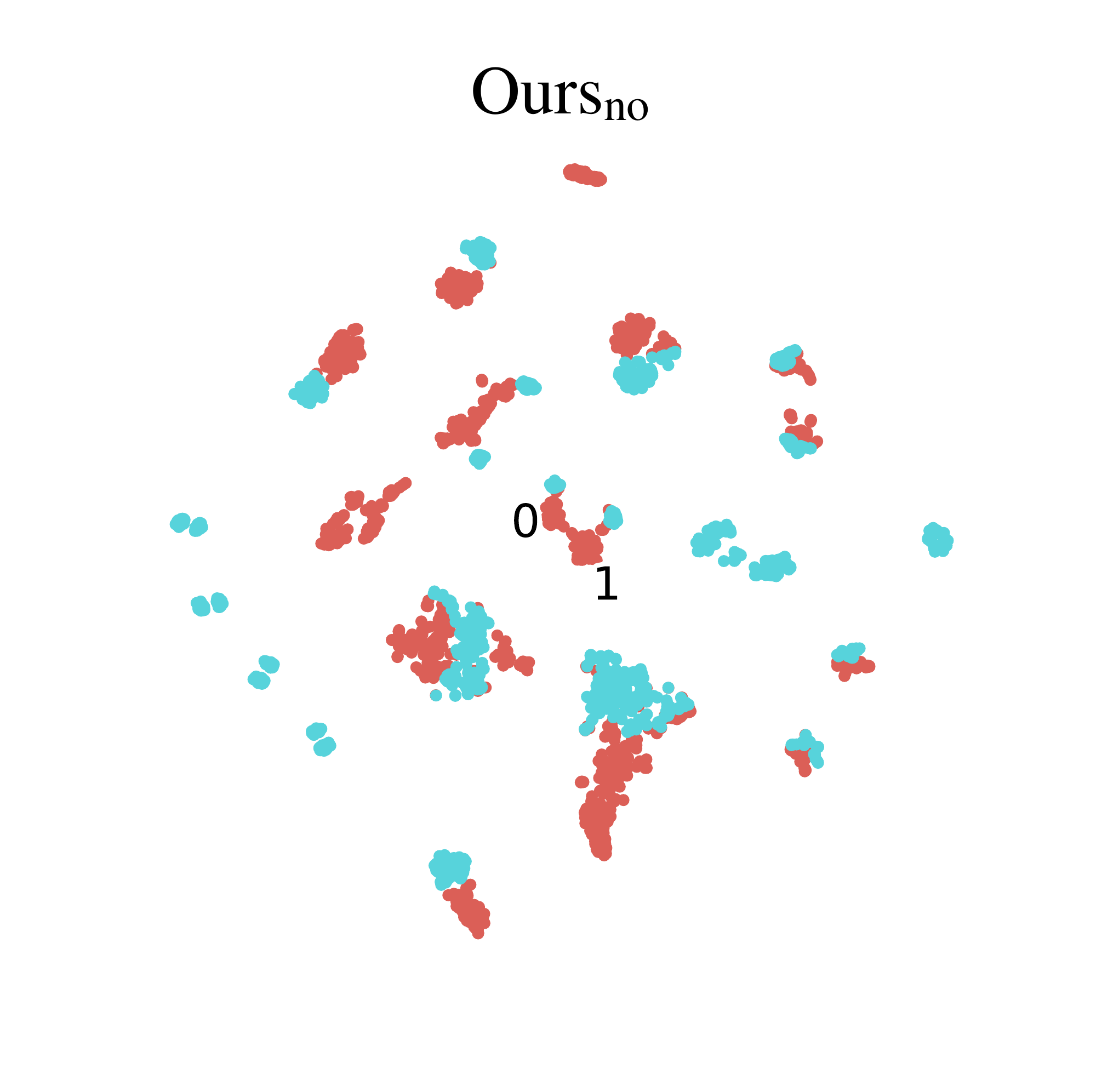}}
	\subfigure[]{
        \includegraphics[width=0.45\linewidth,trim=0 30 0 0]{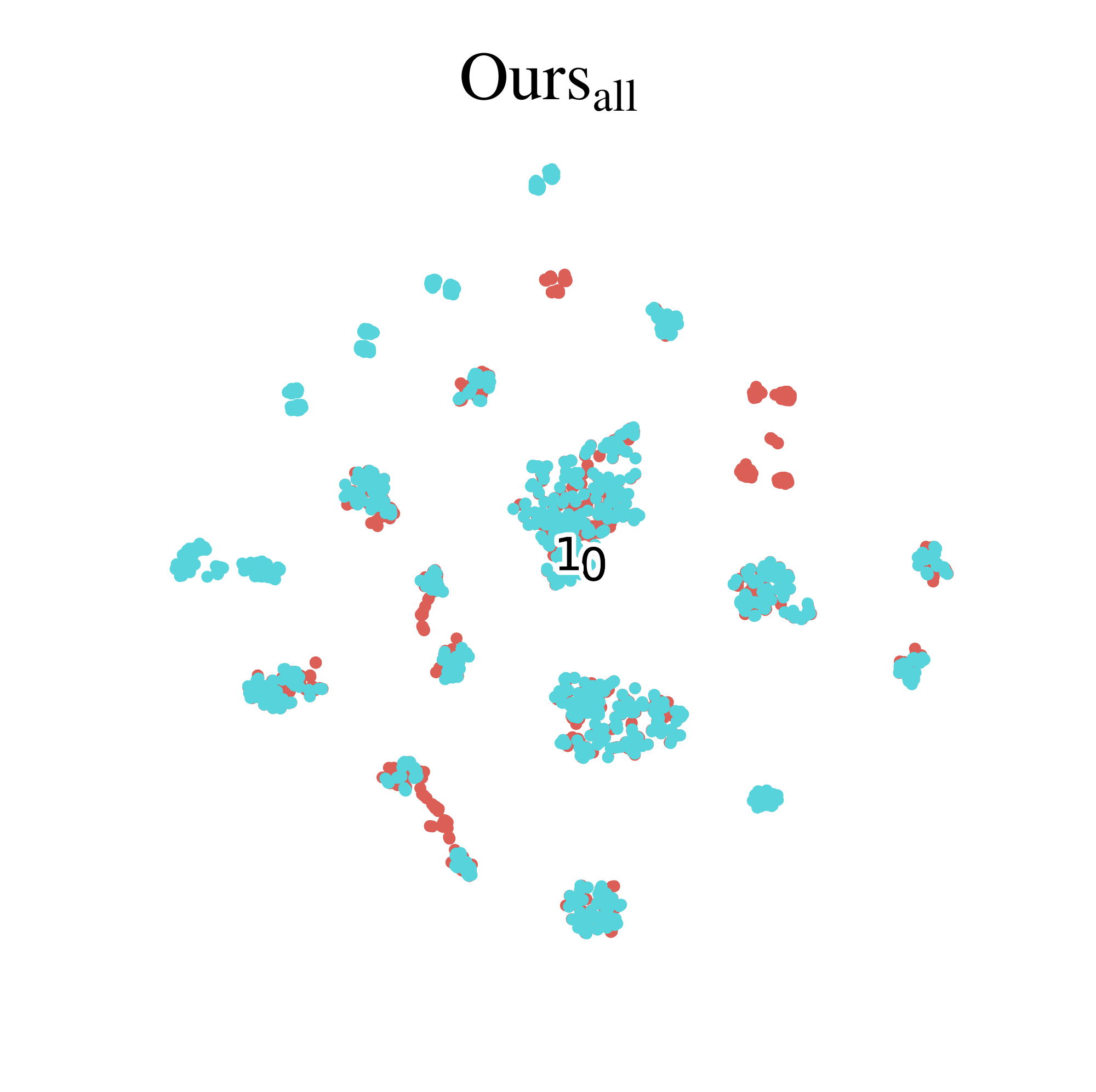}}
	\caption{(a): A t-SNE embedding of the  predicted labels from $\rm Ours_{no}$ and the ground-truth labels on the Multi-PIE database; (b): A t-SNE embedding of the  predicted labels from $\rm Ours_{all}$ and the ground-truth labels on the Multi-PIE database.}
	\label{fig:tsne-cnn}
\end{figure}

In order to further demonstrate the effectiveness of the proposed method in capturing label-level dependencies, we employ t-SNE to depict the predicted label distribution from  $\rm Ours_{no}$,  $\rm Ours_{all}$ and the ground-truth label distributions as shown in Fig.~\ref{fig:tsne-cnn}. Fig.~\ref{fig:tsne-cnn} (a) shows that there exists gap between the predicted label distribution from  $\rm Ours_{no}$ and the ground-truth label distribution.  While in Fig.~\ref{fig:tsne-cnn} (b), the predicted label distribution from  $\rm Ours_{all}$ and the ground-truth label distribution are more closed. It demonstrates that the proposed method successfully captures the ground-truth label distribution through adversarial learning, and thus achieves better performance on multiple face analysis tasks.

\subsection{Experimental results and analyses of multiple face attribute estimation}
The experimental results of face attribute estimation on the CelebA database and LFWA database are listed in Table~\ref{tab:cmp-attr}. From Table~\ref{tab:cmp-attr}, we find that compared to $\rm Ours_{no}$, the proposed method $\rm Ours_{gan}$ achieves 1\% improvement on both databases for average accuracy. For the 40 face attributes, there exist complex correlations. For instance, a person who wears lipstick and necklace is less likely to be a male, while a person with mustache or goatee is more likely to be a male. On the other hand, some face attributes are mutually exclusive. For instance, at most one of black hair, blond hair, brown hair and gray hair appears. These correlations are crucial for improving the performance on multiple attribute estimation simultaneously. The proposed method $\rm Ours_{gan}$ considers the distribution among the predicted face attributes and the ground-truth face attributes. Through this way, the positive correlation and negative correlation among attributes can be exploited. The improvement demonstrates that the proposed method can successfully capture the label-level dependencies and results in better performance.

\begin{figure}[!htbp]
	\centering
    \subfigure[]{
        \includegraphics[width=0.7\linewidth]{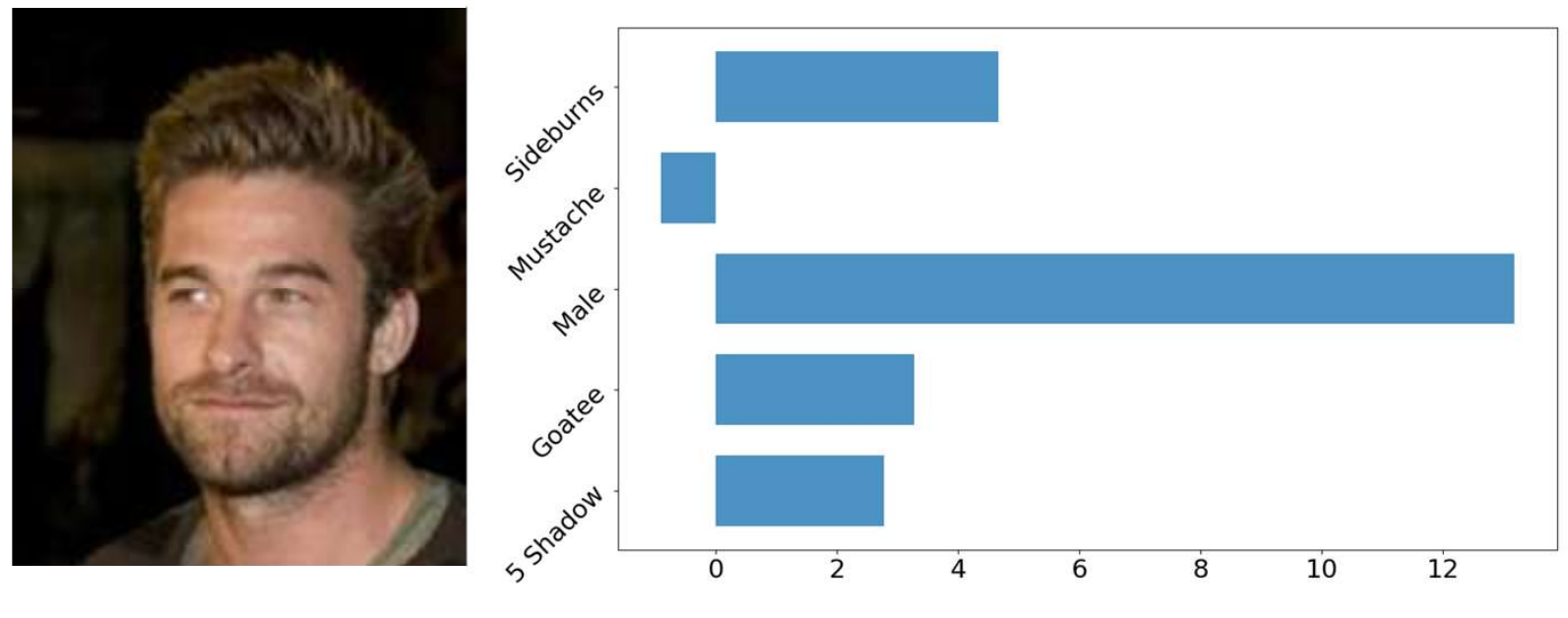}}
    \subfigure[]{
        \includegraphics[width=0.7\linewidth]{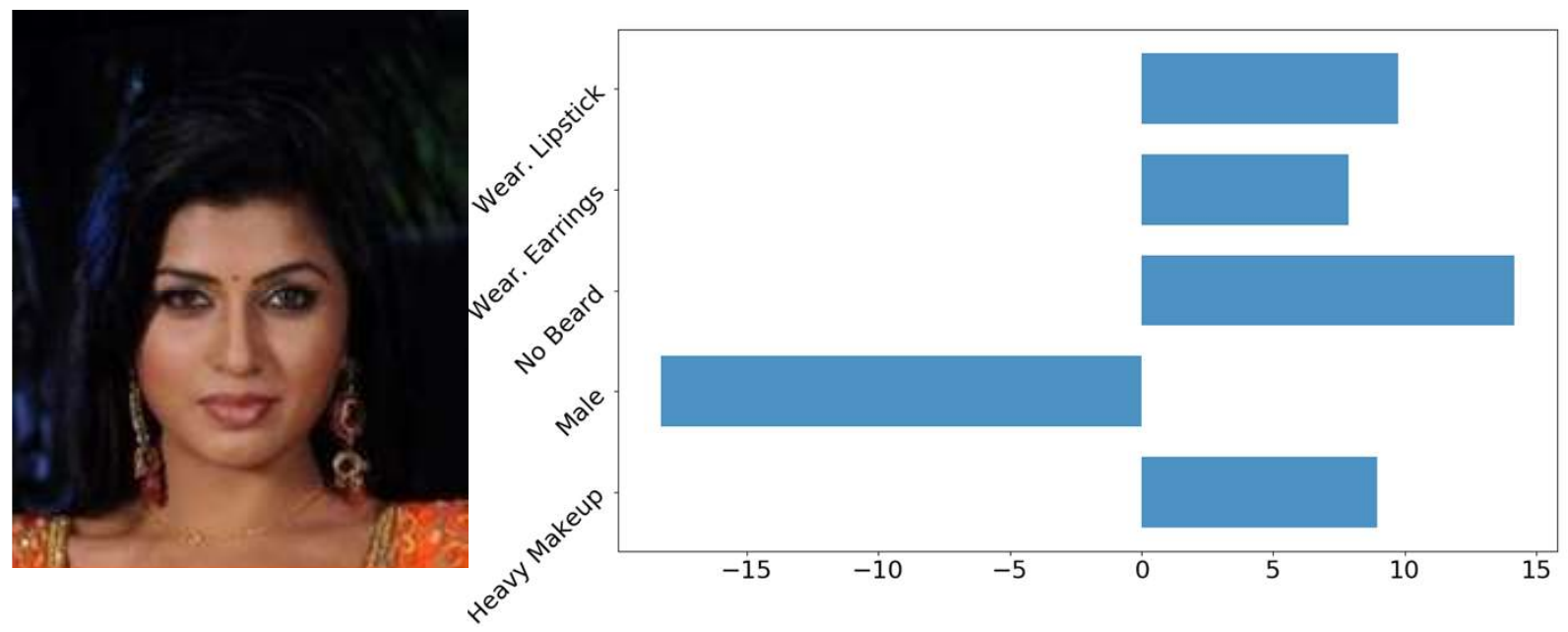}}
	\caption{Example showing: some face attribute combinations are frequently observed. Each bar shows the output value of the recognizer.}
	\label{fig:attr}
\end{figure}

To validate the effectiveness of the proposed method in capturing relationships among multiple face attributes, we graphically illustrate the captured dependencies in Fig.~\ref{fig:attr}. The values are the output of the last layer of the proposed multi-task recognizer $\mathcal{R}$. A larger output value indicates a high confidence of the occurrence for the attribute, and a smaller output value indicates a high confidence of the absence for the attribute. The first figure shows that the sample, which encodes a pattern for a person who is likely to be with sideburns, goatee, 5 o'clock shadow and without mustache. This combination is more likely to represent the attribute relationships for a male. The second figure shows that the sample, which encodes a pattern for a person who is likely to wear lipstick and earrings and with no beard. This combination is more likely to represent the attribute relationships for a heavy makeup female. The two figures show that the proposed method is able to effectively capture the relationships among multiple face attributes.

\begin{table}[!htbp]
	\centering
    \caption{Comparison between the proposed method and the related works of landmark detection (NME(\%)) on the AFLW database (21pts).}
	\label{tab:cmp-aflw-pts}
	\begin{tabular}{l| ccc cc}
		\toprule
		Method & [0, 30] & [30, 60] & [60, 90] & mean & std \\
		\midrule
		CDM~\cite{yu2013pose} & 8.15 & 13.02 & 16.17 & 12.44 & 4.04 \\
		RCPR~\cite{burgos2013robust} & 5.43 & 6.58 & 11.53 & 7.85 & 3.24 \\
		ESR~\cite{cao2014face} & 5.66 & 7.12 & 11.94 & 8.24 & 3.29 \\
		SDM~\cite{xiong2013supervised} & 4.75 & 5.55 & 9.34 & 6.55 & 2.45 \\
		3DDFA~\cite{zhu2016face} & 5.00 & 5.06 & 6.74 & 5.60 & 0.99 \\
		3DDFA+SDM~\cite{zhu2016face} & 4.75 & 4.83 & 6.38 & 5.32 & 0.92 \\
		HyperFace~\cite{ranjan2017hyperface} & 3.93 & 4.14 & 4.71 & 4.26 & 0.41 \\
        Zhang~\emph{et al.}~\cite{zhang2018face} & 3.90 & 4.10 & 4.70 & 4.24 & - \\
		$\rm Ours_{all}$ & \textbf{3.19} & \textbf{3.28} & \textbf{3.49} & \textbf{3.32} & \textbf{0.18} \\
		\bottomrule
	\end{tabular}
\end{table}

\begin{table}[!htbp]
	\centering
    \tabcolsep=4pt
	\caption{Comparison between our method and the related works of landmark detection (MAE) on the Multi-PIE database.}
	\label{tab:cmp-multipie-pts}
	\begin{tabular}{cccccc|cc}
		\toprule
		\multicolumn{6}{c}{near-frontal} & \multicolumn{2}{c}{all poses} \\
		\midrule
		CLM~\cite{saragih2011deformable} & FPLL~\cite{zhu2012face} & CDM~\cite{yu2013pose} & 3DDFA~\cite{zhu2016face} & 3D CLM~\cite{baltruvsaitis20123d} & Chehra~\cite{asthana2014incremental} & Wu~\textit{et al.}~\cite{wu2017simultaneous} & $\rm Ours_{all}$ \\
		4.75 & 4.39 & 7.34 & 5.74 & 5.30 & 4.09 & 3.50 & \textbf{2.85} \\
		\bottomrule
	\end{tabular}
\end{table}

\begin{minipage}[!htbp]{\linewidth}
  \begin{minipage}[htb]{0.42\linewidth}
    \centering
    \makeatletter\def\@captype{table}\makeatother\caption{\footnotesize Comparison between the proposed method and the related works of pose estimation (absolute degree) on the AFLW database.}\label{tab:cmp-aflw-pose}
    \begin{tabular}{l| ccc}
		\toprule
		Method & roll & pitch & yaw \\
		\midrule
		HyperFace~\cite{ranjan2017hyperface} & 3.92 & 6.13 & 7.62 \\
		$\rm Ours_{all}$ & \textbf{3.50} & \textbf{3.08} & \textbf{4.81} \\
		\bottomrule
	\end{tabular}
  \end{minipage}
  \quad
  \begin{minipage}[htb]{0.42\linewidth}
    \centering
    \tabcolsep=2pt
    \makeatletter\def\@captype{table}\makeatother\caption{Comparison between the proposed method and the related works of pose estimation (Acc) on the Multi-PIE database.}\label{tab:cmp-multipie-pose}
    \begin{tabular}{l cccccc}
		\toprule
		& {\footnotesize PCR} & {\footnotesize Linear PLS} & KPLS & {\footnotesize Wu~\textit{et al.}} & {\footnotesize FPLL} & $\rm Ours_{all}$ \\
        & \cite{al2012partial} & \cite{al2012partial} & \cite{al2012partial} & \cite{wu2017simultaneous} & \cite{zhu2012face} & \\
		\midrule
		accuracy & 0.48 & 0.57 & 0.79 & 0.77 & 0.91 & \textbf{0.99} \\
		\bottomrule
	\end{tabular}
  \end{minipage}
\end{minipage}

\begin{table}[!htbp]
  \centering
  \tabcolsep=1.5pt
  \setlength{\abovecaptionskip}{1pt}
  \setlength{\belowcaptionskip}{-1pt}
  \caption{Comparison of Attribution Estimation on the CelebA and LFWA databases.}\label{tab:cmp-attr}
  \begin{tabular}{|c|l|c|c|c|c|c|c|c|c|c|c|c|c|c|c|c|c|c|c|c|c|c|}
    \hline
    Database &  & \rotatebox{90}{5 Shadow} & \rotatebox{90}{Arch. Eyebrows} & \rotatebox{90}{Attractive} & \rotatebox{90}{Bags Un. Eyes} & \rotatebox{90}{Bald} & \rotatebox{90}{Bangs} & \rotatebox{90}{Big Lips} & \rotatebox{90}{Big Nose} & \rotatebox{90}{Black Hair} & \rotatebox{90}{Blond Hair} & \rotatebox{90}{Blurry} & \rotatebox{90}{Brown Hair} & \rotatebox{90}{Bushy Eyebrows} & \rotatebox{90}{Chubby} & \rotatebox{90}{Double Chin} & \rotatebox{90}{Eyeglasses} & \rotatebox{90}{Goatee} & \rotatebox{90}{Gray Hair} & \rotatebox{90}{Heavy Makeup} & \rotatebox{90}{H. Checkbones} & \rotatebox{90}{Male} \\
    \hline
    \multirow{9}{*}{CelebA} & FaceTracker~\cite{kumar2008facetracer} & 85 & 76 & 78 & 76 & 89 & 88 & 64 & 74 & 70 & 80 & 81 & 60 & 80 & 86 & 88 & 98 & 93 & 90 & 85 & 84 & 91 \\
    & PANDA~\cite{zhang2014panda} & 88 & 78 & 81 & 79 & 96 & 92 & 67 & 75 & 85 & 93 & 86 & 77 & 86 & 86 & 88 & 98 & 93 & 94 & 90 & 86 & 97 \\
    & LNets+ANet~\cite{liu2015deep} & 91 & 79 & 81 & 79 & 98 & 95 & 68 & 78 & 88 & 95 & 84 & 80 & 90 & 91 & 92 & 99 & 95 & 97 & 90 & 87 & 98 \\
    & CTS-CNN~\cite{zhong2016face} & 89 & 83 & 82 & 79 & 96 & 94 & 70 & 79 & 87 & 93 & 87 & 79 & 87 & 88 & 89 & 99 & 94 & 95 & 91 & 87 & \textbf{99} \\
    & MCNN-AUX~\cite{hand2017attributes} & 95 & 83 & 83 & 85 & \textbf{99} & 96 & 71 & 85 & 90 & 96 & 96 & 89 & 93 & 96 & 96 & 90 & \textbf{100} & 97 & \textbf{98} & \textbf{99} & 92 \\
    & PS-MCNN-LC~\cite{cao2018partially} & \textbf{97} & \textbf{86} & 84 & 87 & \textbf{99} & 98 & 73 & 86 & 92 & \textbf{98} & 98 & 91 & \textbf{95} & \textbf{98} & \textbf{98} & \textbf{100} & 98 & \textbf{99} & 93 & 90 & \textbf{99} \\
    & Han~\emph{et al.}~\cite{han2018heterogeneous} & 95 & \textbf{86} & \textbf{92} & 85 & 85 & \textbf{99} & \textbf{99} & \textbf{91} & \textbf{96} & 88 & 98 & \textbf{96} & 85 & 96 & 96 & 97 & 99 & \textbf{99} & 92 & 88 & 98 \\
    & $\rm Ours_{no}$ & 95 & 84 & 84 & 87 & \textbf{99} & 97 & 88 & 88 & 91 & 97 & 98 & 90 & \textbf{95} & 96 & 97 & \textbf{100} & 98 & 98 & 93 & 91 & 98 \\
    & $\rm Ours_{gan}$ & 95 & 85 & 86 & \textbf{89} & \textbf{99} & 97 & 92 & 90 & 92 & 97 & \textbf{99} & 91 & \textbf{95} & \textbf{98} & \textbf{98} & \textbf{100} & 98 & \textbf{99} & 93 & 92 & 98 \\
    \hline
    \multirow{9}{*}{LFWA} & FaceTracker~\cite{kumar2008facetracer} & 70 & 67 & 71 & 65 & 77 & 72 & 68 & 73 & 76 & 88 & 73 & 62 & 67 & 67 & 70 & 90 & 69 & 78 & 88 & 77 & 84 \\
    & PANDA~\cite{zhang2014panda} & 84 & 79 & 81 & 80 & 84 & 84 & 73 & 79 & 87 & 94 & 74 & 74 & 79 & 69 & 75 & 89 & 75 & 81 & 93 & 86 & 92 \\
    & LNets+ANet~\cite{liu2015deep} & 84 & 82 & 83 & 83 & 88 & 88 & 75 & 81 & 90 & 97 & 74 & 77 & 82 & 73 & 78 & \textbf{95} & 78 & 84 & 95 & 88 & 94 \\
    & CTS-CNN~\cite{zhong2016face} & 77 & 83 & 79 & 83 & 91 & 91 & 78 & 83 & 90 & 97 & 88 & 76 & 83 & 75 & 80 & 91 & 83 & 87 & 95 & 88 & 94 \\
    & MCNN-AUX~\cite{hand2017attributes} & 77 & 82 & 80 & 83 & 92 & 90 & 79 & 85 & 93 & 97 & 85 & 81 & 85 & 77 & 82 & \textbf{95} & 91 & 83 & 89 & 90 & \textbf{96} \\
    & PS-MCNN-LC~\cite{cao2018partially} & 78 & 84 & 82 & \textbf{87} & 93 & 91 & 83 & 86 & 93 & \textbf{99} & 87 & \textbf{82} & 86 & 78 & \textbf{87} & 93 & 84 & 91 & \textbf{97} & 89 & 95 \\
    & Han~\emph{et al.}~\cite{han2018heterogeneous} & 80 & 86 & 80 & 82 & 84 & 92 & \textbf{93} & \textbf{92} & \textbf{97} & 81 & 88 & 77 & 83 & \textbf{89} & 75 & 78 & \textbf{92} & 86 & 95 & 89 & 93 \\
    & $\rm Ours_{no}$ & 89 & \textbf{88} & \textbf{84} & 82 & \textbf{96} & \textbf{94} & 76 & 83 & 94 & \textbf{99} & 97 & 78 & 80 & 76 & 80 & 92 & 88 & 92 & \textbf{97} & 90 & 93 \\
    & $\rm Ours_{gan}$ & \textbf{90} & \textbf{88} & \textbf{84} & 84 & \textbf{96} & \textbf{94} & 78 & 85 & 95 & \textbf{99} & \textbf{98} & \textbf{82} & \textbf{88} & 81 & 85 & 92 & 90 & \textbf{93} & \textbf{97} & \textbf{91} & 94 \\
    \hline
    & & \rotatebox{90}{Mouth S. O.} & \rotatebox{90}{Mustache} & \rotatebox{90}{Narrow Eyes} & \rotatebox{90}{No Beard} & \rotatebox{90}{Oval Face} & \rotatebox{90}{Pale Skin} & \rotatebox{90}{Pointy Nose} & \rotatebox{90}{Reced. Hairline} & \rotatebox{90}{Rosy Cheeks} & \rotatebox{90}{Sideburns} & \rotatebox{90}{Smiling} & \rotatebox{90}{Straight Hair} & \rotatebox{90}{Wavy Hair} & \rotatebox{90}{Wear. Earrings} & \rotatebox{90}{Wear. Hat} & \rotatebox{90}{Wear. Lipstick} & \rotatebox{90}{Wear. Necklace} & \rotatebox{90}{Wear. Necktie} & \rotatebox{90}{Young} & & \rotatebox{90}{\textbf{Average}} \\
    \hline
    \multirow{9}{*}{CelebA} & FaceTracker~\cite{kumar2008facetracer} & 87 & 91 & 82 & 90 & 64 & 83 & 68 & 76 & 84 & 94 & 89 & 63 & 73 & 73 & 89 & 89 & 68 & 86 & 80 & & 81 \\
    & PANDA~\cite{zhang2014panda} & 93 & 93 & 84 & 93 & 65 & 91 & 71 & 85 & 87 & 93 & 92 & 69 & 77 & 78 & 96 & 93 & 67 & 91 & 84 & & 85 \\
    & LNets+ANet~\cite{liu2015deep} & 92 & 95 & 81 & 95 & 66 & 91 & 72 & 89 & 90 & 96 & 92 & 73 & 80 & 82 & \textbf{99} & 93 & 71 & 93 & 87 & & 87 \\
    & CTS-CNN~\cite{zhong2016face} & 92 & 93 & 78 & 94 & 67 & 85 & 73 & 87 & 88 & 95 & 92 & 73 & 79 & 82 & 96 & 93 & 73 & 91 & 86 & & 87 \\
    & MCNN-AUX~\cite{hand2017attributes} & 88 & 94 & \textbf{98} & 94 & \textbf{97} & 87 & \textbf{87} & \textbf{97} & 96 & 76 & \textbf{97} & 77 & \textbf{94} & \textbf{95} & 98 & 93 & 84 & 84 & 88 & & 91 \\
    & PS-MCNN-LC~\cite{cao2018partially} & \textbf{96} & \textbf{99} & 89 & \textbf{98} & 77 & \textbf{99} & 79 & 96 & \textbf{97} & \textbf{98} & 95 & 86 & 86 & 93 & \textbf{99} & \textbf{96} & 89 & \textbf{99} & \textbf{91} & & 93 \\
    & Han~\emph{et al.}~\cite{han2018heterogeneous} & 94 & 97 & 90 & 97 & 78 & 97 & 78 & 94 & 96 & \textbf{98} & 94 & 85 & 87 & 91 & \textbf{99} & 93 & 89 & 97 & 90 & & 93 \\
    & $\rm Ours_{no}$ & 95 & 98 & 96 & 95 & 83 & \textbf{99} & 82 & 96 & 96 & \textbf{98} & 95 & 84 & 88 & 91 & \textbf{99} & 94 & 94 & 98 & 87 & & 93 \\
    & $\rm Ours_{gan}$ & 95 & 98 & 97 & 95 & 86 & \textbf{99} & 85 & 96 & \textbf{97} & \textbf{98} & 95 & \textbf{91} & 87 & 92 & \textbf{99} & 95 & \textbf{98} & 98 & 90 & & \textbf{94} \\
    \hline
    \multirow{9}{*}{LFWA} & FaceTracker~\cite{kumar2008facetracer} & 77 & 83 & 73 & 69 & 66 & 70 & 74 & 63 & 70 & 71 & 78 & 67 & 62 & 88 & 75 & 87 & 81 & 71 & 80 & & 74 \\
    & PANDA~\cite{zhang2014panda} & 78 & 87 & 73 & 75 & 72 & 84 & 76 & 84 & 73 & 76 & 89 & 73 & 75 & 92 & 82 & 93 & 86 & 79 & 82 & & 81 \\
    & LNets+ANet~\cite{liu2015deep} & 82 & 92 & 81 & 79 & 74 & 84 & 80 & 85 & 78 & 77 & 91 & 76 & 76 & 94 & 88 & 95 & 88 & 79 & 86 & & 84 \\
    & CTS-CNN~\cite{zhong2016face} & 81 & 94 & 81 & 80 & 75 & 73 & 83 & 86 & 82 & 82 & 90 & 77 & 77 & 94 & 90 & 95 & 90 & 81 & 86 & & 85 \\
    & MCNN-AUX~\cite{hand2017attributes} & \textbf{88} & 95 & \textbf{94} & \textbf{84} & \textbf{93} & 83 & \textbf{90} & 81 & 82 & 77 & \textbf{93} & \textbf{84} & \textbf{86} & 88 & 83 & 92 & 79 & 82 & 86 & & 86 \\
    & PS-MCNN-LC~\cite{cao2018partially} & 85 & 94 & 84 & 82 & 78 & \textbf{95} & 88 & 88 & 89 & 84 & \textbf{93} & 80 & 83 & \textbf{96} & 91 & 96 & 91 & 82 & 87 & & 88 \\
    & Han~\emph{et al.}~\cite{han2018heterogeneous} & 86 & 95 & 82 & 81 & 75 & 91 & 84 & 85 & 86 & 80 & 92 & 79 & 80 & 94 & 92 & 93 & 91 & 81 & 87 & & 86 \\
    & $\rm Ours_{no}$ & 83 & 96 & 69 & 73 & 74 & 83 & 80 & 87 & 96 & 84 & 89 & 81 & 80 & 79 & 94 & 97 & \textbf{94} & 82 & 93 & & 87 \\
    & $\rm Ours_{gan}$ & 84 & \textbf{97} & 71 & 80 & 79 & 84 & 83 & \textbf{89} & \textbf{97} & \textbf{87} & 92 & 83 & 83 & 88 & \textbf{95} & \textbf{98} & \textbf{94} & \textbf{84} & \textbf{94} & & \textbf{89} \\
    \hline
  \end{tabular}
\end{table}

\subsection{Comparison with related works on facial landmark related multiple face analyses}

As mentioned in the introduction, several works handle landmark-related multiple face analysis tasks jointly. Among them, Zhu~\emph{et al.}'s work~\cite{zhu2012face} and Wu~\emph{et al.}'s work~\cite{wu2017simultaneous} did not conduct experiments on the AFLW database. Although Zhang~\emph{et al.}'s work~\cite{zhang2014facial} conducted landmark detection experiments on the AFLW database, they adopted mean error as the evaluation metrics, which is different from ours. Therefore, we do not compare our work with Zhu~\emph{et al.}'s, Wu~\emph{et al.}'s and Zhang~\emph{et al.}'s works on the AFLW database. Ranjan~\emph{et al.}'s work~\cite{ranjan2017hyperface} and Zhang~\emph{et al.}'s work~\cite{zhang2018face} provided landmark detection experimental results on the AFLW database with the same experimental conditions with ours. Furthermore, Ranjan~\emph{et al.}'s work~\cite{ranjan2017hyperface} compared their method on landmark detections with CDM~\cite{yu2013pose}, RCPR~\cite{burgos2013robust}, ESR~\cite{cao2014face}, SDM~\cite{xiong2013supervised}, 3DDFA~\cite{zhu2016face} and 3DDFA+SDM~\cite{zhu2016face}. Thus, we compare our work on landmark detection and pose estimation with Ranjan~\emph{et al.}'s work. We also compare our work on landmark detection to CDM, RCPR, ESR, SDM, 3DDFA and 3DDFA+SDM, whose experimental results are directly adopted from Ranjan~\emph{et al.}'s work.

For the Multi-PIE database, we compare our method with Wu~\emph{et al.}'s work~\cite{wu2017simultaneous} and Zhu~\emph{et al.}'s work~\cite{zhu2012face}. Wu~\emph{et al.} provided landmark detection and pose estimation results on the Multi-PIE database with the same experimental conditions with ours. Zhu~\emph{et al.} just evaluated performance on frontal faces. Furthermore, we compare our method with CLM~\cite{saragih2011deformable}, CDM~\cite{yu2013pose}, 3DDFA~\cite{zhu2016face}, 3D CLM~\cite{baltruvsaitis20123d} and Chehra~\cite{asthana2014incremental} for landmark detection and PCR, Linear PLS and KPLS~\cite{al2012partial} for pose estimation, which are compared in Wu~\emph{et al.}'s work.  Ranjan~\emph{et al.}'s work~\cite{ranjan2017hyperface} and Zhang~\emph{et al.}'s~\cite{zhang2014facial} work do not conduct experiments on the Multi-PIE database. Therefore, we can not compare our method with their works on the Multi-PIE database.

In addition, several recent works achieve state of the art landmark detection performance on the AFLW database , i.e, Honari~\emph{et al.}~\cite{honari2018improving}, Dong~\emph{et al.}~\cite{dong2018style} and Dong~\emph{et al.}~\cite{dong2018supervision}. Since they used the revised annotation of the  AFLW database from Zhu~\emph{et al.}~\cite{zhu2015face}, which drop the landmarks of two ears, we do not compare with them.  We compare our work with Zhang~\emph{et al.}~\cite{zhang2018face}, whose experimental conditions are as the same as ours.

The comparison of the proposed method to the related works is shown in Table~\ref{tab:cmp-aflw-pts}-\ref{tab:cmp-multipie-pose}. From these tables, we find that the proposed method performs the best on the AFLW database and the Multi-PIE database for landmark detection and pose estimation. As mentioned in Section~\ref{sec:related-work}, FPLL~\cite{zhu2012face} considered the dependencies from the label-level through topological structure of facial landmark, but ignored the task dependencies inherent in representation-level. Wu~\emph{et al.}~\cite{wu2017simultaneous} proposed an iterative cascade method for simultaneously facial landmark detection, pose and deformation estimation, thus the errors may be inevitable during cascade iteration. Ranjan~\emph{et al.}~\cite{ranjan2017hyperface} proposed a deep multi-task  CNN to exploit the shared representation among different tasks, but failed to capture the label-level constraint. Zhang~\emph{et al.}~\cite{zhang2018face} proposed a 3D reconstruction to detect 2D facial landmark, but ignored the label-level relations. Other works merely considered landmark detection task, ignoring task dependencies.  While the proposed method simultaneously exploits the shared representation and label-level constraint through multi-task network and adversarial mechanism, and thus achieves the best performance.

\subsection{Comparison with related works on multiple face attribute estimation}
For multiple face attribute estimation, we compare the proposed method with FaceTracer~\cite{kumar2008facetracer}, PANDA~\cite{zhang2014panda}, LNets+ANet~\cite{liu2015deep}, CTS-CNN~\cite{zhong2016face}, MCNN-AUX~\cite{hand2017attributes}, PS-MCNN-LC~\cite{cao2018partially} and Han~\emph{et al.}~\cite{han2018heterogeneous}. The results are listed in  Table~\ref{tab:cmp-attr}.

The proposed method achieves higher average accuracy on both databases than FaceTracker~\cite{kumar2008facetracer}. Specifically, the average accuracy of the proposed method is higher than that achieved by FaceTracker by 13\% on the CelebA database and 12\% on the LFWA database. FaceTracker handled multiple attributes separately, and ignored the dependencies among attributes, which are crucial for estimating multiple face attributes. Therefore, the proposed method outperforms them and achieves higher average accuracy.
Compared to PANDA~\cite{zhang2014panda}, LNets+ANet~\cite{liu2015deep} and CTS-CNN~\cite{zhong2016face}, which estimated multiple attributes through deep convolutional neural network, the proposed method also performs better. Specifically, the average accuracy of the proposed method is 9\%, 7\% and 7\% higher than PANDA, LNets+ANet and CTS-CNN respectively on the CelebA database, and 8\%, 5\% and 4\% higher than PANDA, LNets+ANet and CTS-CNN respectively on the LFWA database. For the three methods, multiple face attributes share common representation, and thus the dependencies among attributes can be exploited in a certain. However, the dependencies on label-level are not considered. Compared to these works, the proposed method considers the representation-level dependencies and the label-level dependencies jointly, and thus achieves better experimental results.
The proposed method outperforms MCNN-AUX~\cite{hand2017attributes}, Han~\emph{et al.}~\cite{han2018heterogeneous} and PS-MCNN-LC~\cite{cao2018partially}, which exploited both the representation-level and label-level dependencies. Specifically, the average accuracy of the proposed method is 3\%, 1\% and 1\% higher than MCNN-AUX, Han~\emph{et al.} and PS-MCNN-LC respectively on the CelebA database, and 3\%, 3\% and 1\% higher than MCNN-AUX, Han~\emph{et al.} and PS-MCNN-LC respectively on the LFWA database. MCNN-AUX proposed an auxiliary network to obtain relationships among multiple face attributes. PS-MCNN-LC and Han~\emph{et al.} grouped these multiple face attributes according to prior knowledge. Although these works exploited dependencies among multiple face attributes from both representation-level and label-level, the captured label-level relationships are either fixed groups or fixed form. Through multi-task adversary network, the proposed method can capture the complex and global relationship among multiple face attributes. On both databases, the proposed method achieves the best performance. It further suggests that the proposed method has strong ability for multi-task analyses.

\section{Conclusion}\label{sec:conclusion}
In this paper, we propose a novel multiple facial analysis method through exploiting both representation-level and label-level dependencies. Specifically, we first utilize deep multi-task network as a recognizer $\mathcal{R}$ to capture representation-level dependencies. And then, we introduce a discriminator $\mathcal{D}$ to distinguish the label combinations from the ground-truth. Through optimizing the two networks in an adversarial manner, the proposed method manages to make predicted label combination closer to the distribution of the ground-truth. Experimental results on four databases demonstrate that the proposed method successfully captures both the shared representation-level and label-level constraint and thus outperforms related works.


%

%
\bibliographystyle{ACM-Reference-Format}
\bibliography{bibliography}

\end{document}